\documentclass[letterpaper, 10 pt, conference]{ieeeconf}

\IEEEoverridecommandlockouts                             
\usepackage{cite}
\usepackage{epsfig} 
\usepackage{epstopdf}
\usepackage{amsmath}
\usepackage{bm}
\usepackage{soul}
\usepackage{amsfonts}
\usepackage{color}
\usepackage{listings}
\usepackage{wrapfig}
\usepackage{lscape}
\usepackage{rotating}
\usepackage{dirtytalk}
\usepackage{adjustbox}
\usepackage{units}
\usepackage{xcolor,colortbl}
\definecolor{Gray}{gray}{0.9}
\usepackage{url}
\usepackage{hyperref}   
\usepackage[makeroom]{cancel}
\usepackage{pifont}
\usepackage{multirow}

\usepackage{graphicx}
\usepackage{algorithm}
\usepackage{algorithmic}

\usepackage{mdframed}

\title{Optimization-Based Quadrupedal Hybrid Wheeled-Legged Locomotion}

\author{I. Belli$^{1}$, M. Parigi Polverini$^{2}$, A. Laurenzi$^{2}$, E. Mingo Hoffman$^{2}$, P. Rocco$^{1}$, N. G. Tsagarakis$^{2}$
\thanks{This work was supported by the European  Union’s  Horizon  2020  Research  and Innovation  Program  under  Grant No. 779963 (EUROBENCH).}
\thanks{$^{1}$ I. Belli and P. Rocco are with Dipartimento di Elettronica, Informazione e Bioingegneria (DEIB), Politecnico di Milano, Milan, Italy. Email: {\tt\footnotesize italo.belli@mail.polimi.it, paolo.rocco@polimi.it}}%
\thanks{$^{2}$ M. Parigi Polverini, A. Laurenzi, E. Mingo Hoffman and N. G. Tsagarakis are with the Humanoids \& Human Centered Mechatronics Research Line (HHCM), Istituto Italiano di Tecnologia (IIT), Genova, Italy. Email: {\tt\footnotesize \{matteo.parigi, enrico.mingo, arturo.laurenzi, nikos.tsagarakis\}@iit.it}}%
}

\begin{document}

\maketitle

\begin{abstract}
This paper presents a trajectory optimization approach to the motion generation problem of hybrid locomotion strategies for a wheeled-legged quadrupedal robot with steerable wheels. To this end, traditional Single Rigid Body Dynamics has been employed and extended by adding a unicycle model for each leg, conveniently incorporating the nonholonomic rolling constraints. The proposed approach can generate hybrid locomotion strategies as well as pure driving and legged locomotion with minimum effort for the user. The effectiveness of the proposed approach has been experimentally validated on the humanoid quadruped CENTAURO, employing a hierarchical inverse kinematics engine to track the planned motions. 
\end{abstract}

\section{Introduction} \label{sec:introduction}
\begin{figure}
    \centering
    \includegraphics[width=.95\columnwidth]{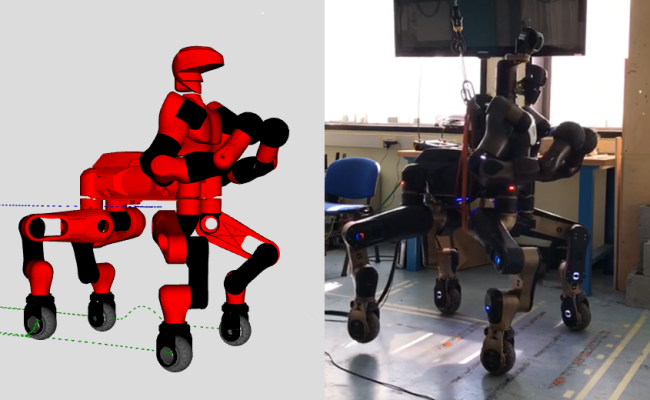}
    \caption{RViz visualization and experimental validation on the robotic platform CENTAURO of a hybrid locomotion trajectory produced by the proposed approach. In this snapshot the rear right foot is lifted, while wheeled navigation is simultaneously performed.}
    \label{fig: intro}
\end{figure}

Hybrid wheeled-legged locomotion is a navigation paradigm only recently opened up by novel robotic designs, e.g. the centaur-type humanoid CENTAURO \cite{kashiri2019centauro} or the quadruped ANYmal \cite{keeprollin} in its configuration featuring non-steerable wheels. The term \textit{Hybrid Locomotion} is hereafter used to indicate a particular type of locomotion, achieved with simultaneous and coordinate use of legs and wheels, see Fig. \ref{fig: intro}. Such choice stems at the intersection between legged locomotion and the simpler wheeled navigation, in order to get the best from both techniques: agility and ability to traverse uneven terrains from the first, speed and stability from the second. As a consequence, the problem of planning feasible trajectories for a hybrid robot shares many similarities with the legged locomotion problem: also in the hybrid case the motion of the base is reached through contact of the feet with the environment, taking into account that the wheeled feet can just push on the ground and not pull it. Forces compatible with friction cones have to be considered, while the contacts can slide just along the direction prescribed by the orientation of the wheels. 

\subsection{Related works}
Despite the possibilities unfolded by hybrid wheeled-legged robots, the research community has started to explore this field only in very recent years, thus comparatively few works can be found in the literature.
The works in \cite{justin, suzumura2013real, GRAND2010477} focus on hybrid robots where the legs are used just for active suspension during statically-stable driving motions. The work in \cite{drcHubo} presented one of the first legged robots that could exploit wheels to locomote: DRC-HUBO+. With this platform the two types of locomotion are achievable in different configurations of the robot (erect or crunched), hence limiting the motion possibilities. This is also the disadvantage of RoboSimian platform \cite{robosimian}. The centaur-like robot \textit{Momaro} \cite{momaro} instead  has proved to be capable of hybrid locomotion and motion planning, achieved through A$^{*}$-search on a pose grid and relying on height maps of the environment. The recent works in \cite{keeprollin, rollingInTheDeep} address Trajectory Optimization (TO) for the quadrupedal robot ANYmal \cite{anymal} featuring it with actuated non-steerable wheels and taking into account the whole-body dynamic model, wheels included. While in \cite{keeprollin} the robot still performs switches between driving and stepping, in \cite{rollingInTheDeep} coordinate movements are achieved with an online TO framework capable of running in a Model Predictive Control (MPC) fashion, by breaking down the problem into separate wheel and base optimizations. While being computationally efficient, this approach decouples two inherently related quantities (base and feet positions) whose evolution is instead strictly linked. A second issue with this approach is related to the contact forces, that are not decision variables of the problem and hence are not guaranteed to be dynamically consistent. This is solved by the very recent work in \cite{bjelonic2020whole}, that treats the hybrid locomotion problem altogether in a single task. Another interesting work is \cite{pholus}, featuring the robot Pholus (a robotic twin of CENTAURO), that addresses a topic very similar to the one we present in this paper. 
While being able to produce different hybrid movements on the real robot, this approach too suffers from the fact that the optimization of the foothold positions is performed before the one for the center of mass (CoM). In addition, the angular momentum is ignored and the whole trajectory of the CoM along the z axis is assumed to be known in advance, thus greatly limiting the variety of the motions produced and preventing the optimization to run in more general environment.
For completeness, note that similar TO considerations pertain also to pure quadrupedal legged locomotion, a field in which impressive results have been achieved, e.g. \cite{kalakrishnan2011learning}, \cite{wink1stpaper}, \cite{wink3rdpaper}.
\subsection{Contributions}
The main contributions of this paper are listed in the following:
\begin{itemize}
    \item[-] TO is formulated offline for the locomotion problem of a quadrupedal robot with steerable wheels, which is able to produce hybrid gaits as well as pure driving and legged locomotion with minimum effort for the user; 
    \item[-] Single Rigid Body Dynamics (SRBD) is employed, aiming at meeting computational efficiency, and combined with a unicycle model for each leg, in order to account for the presence of steerable wheels;
    \item[-] A detailed description of the constraints to which the system is subject to, including pure rolling conditions, is presented, together with the suitable cost terms that guarantee smooth motion profiles;
    \item[-] Execution of the planned motion, by feeding the optimal trajectories to a Hierarchical Inverse Kinematic (HIK) engine, is experimentally performed on CENTAURO.
\end{itemize}

\section{Modelling of Quadruped Hybrid System} \label{sec:modelling}
\subsection{Robot Dynamic Model}
Reasoning about the full hybrid dynamics of a legged system remains computationally expensive for high-dimensional systems. Reduced order models, such as centroidal dynamics \cite{orin2013centroidal} or SRBD \cite{wink3rdpaper} (the approach that will be adopted in this paper), conveniently offer the possibility to reduce computational complexity and to reason in Cartesian coordinates, although preventing to account for joint limits.
The ordinary differential equations (ODEs) describing the SRBD evolution of the system are:
\\
\begin{subequations} \label{math: srbd_theory}
\begin{alignat}{4}
    & m \boldsymbol{\ddot{r}}=m\boldsymbol{g}+\sum_{i=1}^{4}\boldsymbol{f}_{C,i} \label{math: srbd1} \\
    & \boldsymbol{I}\boldsymbol{\dot{\omega}}+\boldsymbol{\omega} \times \boldsymbol{I} \boldsymbol{\omega} = \sum_{i=1}^{4}\boldsymbol{f}_{C,i} \times (\boldsymbol{r}-\boldsymbol{p}_{C,i}) \label{math: srbd2}
\end{alignat}
\end{subequations}
\\
where \eqref{math: srbd1} is concerned with the evolution of the CoM coordinate \mbox{$\boldsymbol{r} \in \mathbb{R}^3$} and  \eqref{math: srbd2} describes how the angular momentum changes as function of the contact points and contact forces. \eqref{math: srbd1} involves $m \in \mathbb{R}$, representing the overall mass of the robot, $\boldsymbol{g} \in \mathbb{R}^3$ is the gravitational acceleration vector, and $\boldsymbol{f}_{C,i} \in \mathbb{R}^3$ is the contact force at the $i^{th}$ foot. On the other hand, \eqref{math: srbd2} includes the full-body inertia matrix $\boldsymbol{I}\in \mathbb{R}^{3 \times 3}$, the base angular velocity and acceleration $\boldsymbol{\omega}$,  $\dot{\boldsymbol{\omega}}\in \mathbb{R}^3$ as well as the $i^{th}$ foot Cartesian position $\boldsymbol{p}_{C,i} \in \mathbb{R}^3$.
Adoption of this model 
requires the following assumptions:
the robotic links are rigid, the wheeled feet interact with the ground via point-like contacts, the full-body inertia is constant throughout the motion and, finally, the momentum produced by the joint velocities is negligible.
\subsection{Wheels’ Kinematic Model}
\begin{figure}[!h]
    \centering
    \includegraphics[width=.6\columnwidth]{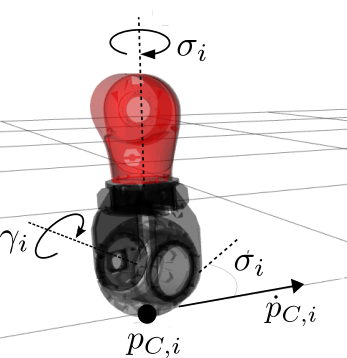}
    \caption{The unicycle kinematic model for each steerable wheel.}
    \label{fig: unicycle}
\end{figure}

The steerable wheels are modelled with a classical unicycle. This means that each foot is completely described by its Cartesian position $\boldsymbol{p}_{C,i}$ and the related steering angle $\sigma_i \in \mathbb R$, see Fig. \ref{fig: unicycle}. 
\\
Pure rolling condition is assumed when the foot is touching flat ground, hence introducing constraints of nonholonomic nature in the system. Considering the $i^{th}$ foot, its velocity along the x-y world axes must be coherent with the wheel steering set by the value of $\sigma_i$:
\begin{equation}
    \dot{p}_{C,i}^x \ sin(\sigma_i)-\dot{p}_{C,i}^y \ cos(\sigma_i)=0
\end{equation}
The presence of this constraint allows the solver to consider just velocities that do not involve lateral slippage of the feet. The rolling speed of the wheel $\dot \gamma_i \in \mathbb R$ can be computed based on the x-velocity of the contact point in the contact frame as:  \mbox{$\dot \gamma_i = ^{\{C,i\}}\dot{p}_{C,i}^x/R$}, being $R$ the wheel radius.

\section{Motion Generation for Hybrid Locomotion}  \label{sec:motion_planning_hybrid}
In the following, we formulate and discuss the Optimal Control Problem (OCP) conceived to drive as desired the evolution of the system described in Sec. \ref{sec:modelling}. Key elements in this process are the definition of the state variables, the control variables, the constraints  and the cost function, that allows one to set the optimality criterion with respect to which the best sequence of values for the control inputs is found. 

\subsection{State Variables}
Given the modelling choice, the hybrid system is fully defined by the following set of variables: \footnote{For ease of notation,the transpose sign of each quantity has been omitted.}
\\
\begin{equation}
    \boldsymbol{x} = \begin{bmatrix} \boldsymbol{r} & \boldsymbol{\theta} & \dot{\boldsymbol{r}} & \dot{\boldsymbol{\theta}} & \boldsymbol{p} \end{bmatrix}^{\top} \in \mathbb{R}^{28}
\end{equation}
where $\boldsymbol{p} \in \mathbb R^{16}$ collects the Cartesian position and steering angle of each wheel, i.e. $\boldsymbol{p}_i=\begin{bmatrix}\boldsymbol{p}_{C,i} & \sigma_i \end{bmatrix}^{\top} \in \mathbb{R}^4$, for \mbox{$i=1,\dots,4$}, while $\boldsymbol{\theta} \in \mathbb{R}^3$ collects the Euler angles describing the orientation of the base of the robot (order of application: yaw-pitch-roll). It has to be noted that, while this definition of the state vector $\boldsymbol{x}$ appears convenient, Eq. (\ref{math: srbd2}) involves angular velocities and accelerations expressed in a fixed world frame, hence a mapping between $\boldsymbol{\theta}$ and $\boldsymbol{\omega}$ is required, also for what concerns the time derivatives of those. Such a mapping is readily provided in \cite{gehring2016kindr}. 
\subsection{Control Inputs}
In our framework, we split the control vector as follows:
\[ \boldsymbol{u}=[\boldsymbol{u}_1 \ \boldsymbol{u}_2]^\top \]
where
\setcounter{MaxMatrixCols}{12}
\begin{subequations} \label{math: control_hybrid}
\begin{alignat}{4}
& \boldsymbol{u}_1=\begin{bmatrix}\boldsymbol{f}_{C} & \boldsymbol{v} & \dot{\boldsymbol{\sigma}} \end{bmatrix}^\top \in \mathbb{R}^{20} \\
& \boldsymbol{u}_2= \dot{\boldsymbol{p}}_C^z \in \mathbb{R}^4
\end{alignat}
\end{subequations} 
Here $\boldsymbol{f}_C \in \mathbb{R}^{12}$ collects the contact forces $\boldsymbol{f}_{C,i}$ of each foot, $\boldsymbol{v} \in \mathbb{R}^4$ collects the scalars describing the velocity of each wheel along its direction of rolling, $\dot{\boldsymbol{\sigma}} \in \mathbb{R}^4$ collects the yaw rate of each wheel, and $\dot{\boldsymbol{p}}_C^z$ collects the vertical component of the speed of each wheeled-foot.
The partition of $\boldsymbol{u}$ into two distinct sub-vectors will become clearer in the following, but can be explained here as necessary to identify the control inputs that directly act on the dynamical evolution of the system (grouped in $\boldsymbol{u}_1$), separating them from the inputs that are used to achieve a good vertical shape of the feet trajectories during the flight phases (namely $\boldsymbol{u}_2$) but do not interfere with the dynamics.

\subsection{Constraints}
To guide the solver towards a feasible solution, some constraints are needed. They contribute to make explicit all the physical limitations proper of CENTAURO that are not dealt with by the aforementioned SRBD model, or are used to select desired behaviours leading to smooth trajectories. In the following we list them, explaining their role.
\begin{itemize}
    \item[-] \textit{Gait sequence}: a classical crawl gait \cite{mcghee1968stability} is used, but a main property of hybrid locomotion is that the goal can be reached also with very different gaits. Even pure driving mode could be used, where the robot just utilizes its wheels without lifting any feet. 
    \item[-] \textit{Contact constraints}: given the time interval $T_{\mathcal{C}}$ in which a foot is in contact with the ground, the following level constraint is needed:
    \begin{equation} \label{math: contactConst_hybrid}
        p_{C,i}^z(t \in T_{\mathcal{C}})=z_{ground}
    \end{equation}
    When foot $i$ is lifted, we make sure that no contact force coming from that specific foot is present, imposing:
    \begin{equation} \label{math: liftedForce_hybrid}
        \boldsymbol{f}_{C,i}(t \not\in T_{\mathcal{C}})=\boldsymbol{0}
    \end{equation}
    By indicating explicitly with $T_{\mathcal{L}}$ the interval in which a foot is lifted, we also shape the vertical trajectory prescribing its maximum height $z_{fly}$:
    \begin{equation} \label{math: liftedFoot_hybrid}
        p_{C,i}^z(\nicefrac{T_{\mathcal{L}}}{2})=z_{fly}
    \end{equation}
    \item[-] \textit{Kinematic constraints}: they link the evolution of the CoM position to the feet positions, hence ensuring feasibility of both the motions together. As the joint limits are not explicitly considered in our model, we translate them into a Cartesian constraint for the relative position of the feet with respect to the CoM. Each foot workspace is approximated with a cuboid of dimensions $\boldsymbol{b} \in \mathbb{R}^3$ that is centered in the nominal position of the foot $\overline{\boldsymbol{p}}_{C,i}$. The constraint for foot $i$ is written as:
        \begin{equation} \label{math: kinConstr}
        \big|\boldsymbol{R}_z(\theta_z)\big(\boldsymbol{p}_{C,i}(t)-\boldsymbol{r}(t)\big)-\overline{\boldsymbol{p}}_{C,i}\big|<\boldsymbol{b} 
    \end{equation}
    Note that in principle taking into account $\boldsymbol{\theta}$ is required, but restricting the rotation to the base yaw angle $\theta_z$ alone is appropriate when focusing on hybrid locomotion, as it generally happens over reasonably flat environments. If other factors, such as the presence of stairs, impose higher pitch or roll angles, one can always resort to the more general $\boldsymbol{R}(\boldsymbol{\theta})$. Another constraint coming from the mechanical structure of the robot involves the wheel steering angles (in world reference frame), bounded to be inside a range of values:
    \begin{equation} \label{math: kinConstr2_hybrid}
        \sigma_{min} \leq \sigma_i-\theta_z \leq \sigma_{max} \qquad \forall i=1,...,4
    \end{equation}
    where $\sigma_{min}$ and $\sigma_{max}$ come from the fact that CENTAURO steering motors are characterized by hard stops.
         \item[-] \textit{Friction cones}: they introduce the physical limitations acting on the contact forces. Indeed, the desired motion for the CoM is achieved by controlling the contact forces between the feet and the ground, coherently with the floating-base model described by (\ref{math: srbd_theory}). Following classical Coulomb friction model, it is necessary that the components of the forces lying on the ground do not exceed a certain threshold, defined by a suitable friction coefficient $\mu$ and by the magnitude of the force projection on an axis that is normal to the ground itself. Even in case of wheeled locomotion, the instantaneous velocity between the feet and the ground is zero (as no slippage is allowed), hence the problem of limiting the values of the forces can be treated as in the legged case. The three components of each force $\boldsymbol{f}_{C,i}$ must be related by the following inequalities:
    \begin{subequations} \label{math: friction_const}
    \begin{alignat}{4}
    & -\mu f_{C,i}^z < f_{C,i}^x < \mu f_{C,i}^z \\
    & -\mu f_{C,i}^z < f_{C,i}^y < \mu f_{C,i}^z
    \end{alignat}
    \end{subequations}
    where the considered axis are those of the contact surface, and we adopt linearized friction cones to avoid time consuming computations while introducing only negligible errors, as in \cite{wink3rdpaper}. In short, we can state that each contact force must belong to a set of admissible values:
    \begin{equation}
        \boldsymbol{f}_{C,i} \in \mathcal F(\bm{f}_{C,i}, \mu) \quad \forall i=1,...,4
    \end{equation}
    \item[-] \textit{Initial and final constraints}: the initial and final value for the state must be given to the algorithm. The initial position, named $\boldsymbol{x}_0$, is a hard equality constraint:
    \begin{equation} \label{math: initial_const}
        \boldsymbol{x}(0)=\boldsymbol{x}_0
    \end{equation}
    while the goal constraint can be suitably enforced in terms of zero final speed (that regards both the CoM and the wheels) as well as final position of the feet:
    \begin{subequations} \label{math: final_const}
    \begin{alignat}{4}
    &\boldsymbol{\dot{r}}(T_f)=\boldsymbol{0} \\
    & v_i(T_f)=0 \\
    & \boldsymbol{R}_z(\theta_z) \big(\boldsymbol{p}_{C,i}(T_f)-\boldsymbol{r}(T_f)\big)=\overline{\boldsymbol{p}}_{C,i} 
    \end{alignat}
    \end{subequations}
    where $T_f$ is the length of the optimization horizon.
    Even if it appears desirable to enforce a constraint directly on the final position of the CoM, this choice would lead the optimization to fail in unforeseen situations. As a consequence, the task of moving the CoM towards the goal position is left to the cost function alone.
\end{itemize}

\subsection{Cost Terms} \label{sec:cost_funct}
The cost function required to achieve the desired task with smooth and well-behaved trajectory is crucial and must be defined carefully. In our framework it is composed of several terms, listed in the following:
\begin{itemize}
    \item[-] $L_{acc}=\boldsymbol{\dot{x}}^ \top \boldsymbol{Q}_1 \boldsymbol{\dot{x}}$, accounting for the system acceleration, whose minimization is required in order to produce smooth motions, to facilitate tracking as well as to reduce joint torques and energy consumption. A positive semidefinite matrix $\boldsymbol{Q_1} \in \mathbb{R}^{28 \times 28}$ allows to consider just the CoM accelerations $\boldsymbol{\ddot{r}}$.
    \item[-] $L_{goal}=(\boldsymbol{x}-\boldsymbol{x}_{goal})^\top \boldsymbol{Q}_2 (\boldsymbol{x}-\boldsymbol{x}_{goal})$, accounting for the distance to the desired goal position and orientation. Again, a positive semidefinite matrix $\boldsymbol{Q_2} \in \mathbb{R}^{28 \times 28}$ is used to select the CoM position vector $\boldsymbol{r}$ and orientation $\boldsymbol{\theta}$. Therefore just the Cartesian and angular distance to $\boldsymbol{x}_{goal} \in \mathbb{R}^{28}$ are weighted, while the rates of variation of these quantities are not considered to leave the solver free to adjust them as required.
    \item[-] $L_{force}=\sum_{i=1}^{4}\boldsymbol{\delta f}_{C,i}$, accounting for the variation of the contact forces over time. Preventing the contact forces from varying too suddenly allows the feet to be able to actually exert the correct force at every time instant. The implementation of this cost term is done, when the continuous OCP is cast into discrete time, by computing the difference between the force at the \mbox{$k$} and \mbox{$k-1$} time intervals.
    \item[-] $L_{feet}=\sum_{i=1}^{4}(\boldsymbol{R}_z(\theta_z)(\boldsymbol{p}_{C,i}-\boldsymbol{r})-\overline{\boldsymbol{p}}_{C,i})^2$, accounting for the distance of each foot from its nominal position $\overline{\boldsymbol{p}}_{C,i}$ with respect to the CoM. The distance is evaluated in the world reference frame and then expressed in the robot base reference frame to compare it with the nominal one. To achieve faster computation, we consider again just the base yaw angle $\theta_z$, similarly to what was done in (\ref{math: kinConstr}).
    \item[-] $L_{speed}=\sum_{i=1}^{4}v_i^2$, accounting for the linear speed of each rolling wheel. It is added, usually with a small weight, for preventing the feet from moving too fast, reducing the torque required for the driving motors, as well as for penalizing the swinging motion of the wheels arising when the robot CoM has already reached its final goal. Interestingly, if considered just for the feet that are in contact with the ground, it represents a powerful tuning element to span between car-like and pure walking behaviours. 
    \item[-] $L_{yaw}=\sum_{i=1}^{4}\dot{\sigma}_i^2$, accounting for the steering speed of each wheel, that is the velocity with which it reorients when required. Intuitively, steering the wheels too fast is not recommendable as it might hinder the stability of the robotic platform, whose inertia is such that sudden variations in the direction of the motion could even lead the robot to fall.
\end{itemize}
A weighted sum of all the terms above builds up the cost function that is used in the current paper. 

\subsection{OCP Formulation}
The overall OCP we address in this Section reads as:
\begin{mdframed}
\begin{equation} 
\label{math: ocp}
\begin{aligned}
& \underset{\bm x(\cdot), \bm u(\cdot)}{\operatorname{min}} \ \ \int_0^{T_f} L\big(\bm x(t),\bm u(t)\big)dt\\
& \text{subject to}\\ 
\end{aligned}
\end{equation}
\vspace{-0.5cm}
\begin{IEEEeqnarray*}{lc}
    \quad \bm{x}(0) - \bm{x}_{0} = \bm{0} & \text{initial state}\\
    \quad \dot{\boldsymbol{r}}(T_f)=\boldsymbol{0} & \text{final velocity}\\
    \quad \forall \ \text{foot} \ i: \\
    \qquad \boldsymbol{R}_z(\theta_z)(\boldsymbol{p}_{C,i}(T_f)-\boldsymbol{r}(T_f))=\overline{\boldsymbol{p}}_{C,i} & \text{final position}\\
    \qquad \big|\boldsymbol{R}_z(\theta_z)(\boldsymbol{p}_{C,i}(t)-\boldsymbol{r}(t))-\overline{\boldsymbol{p}}_{C,i}\big|<\boldsymbol{b} & 
    \text{kin. limits}\\
    \qquad \sigma_{min} \leq \sigma_i(t)-\theta_z(t) \leq \sigma_{max} & \text{kin. limits}\\
    \qquad \dot{\sigma}_{min} \leq \dot{\sigma}_i(t) \leq \dot{\sigma}_{max} & \text{actuator limits}\\
    \qquad v_i(T_f)=0 & \text{final velocity} \\
    \qquad \text{if foot in contact:} \\
    \quad \quad \quad p_{C,i}^z(t \in T_{\mathcal{C}})=z_{ground} & \text{foot level}\\
    \quad \quad \quad \bm{f}_{C,i}(t \in T_{\mathcal{C}} )  \in \mathcal F(\bm{f}_{C,i}, \bm \eta_\mathcal{S}, \mu) \quad & \text{friction cone} \\
    \quad \quad \quad v_{min} \leq v_i(t) \leq v_{max} & \text{actuator limits}\\
    \qquad \text{otherwise:} \\
    \quad \quad \quad \bm{f}_{\text{C,i}}(t \not\in T_{\mathcal{C}}) = \bm 0 \quad & \text{no force} \\
    \quad \quad \quad p_{C,i}^z(\nicefrac{T_{\mathcal{L}}}{2})=z_{fly} & \text{foot level}
\end{IEEEeqnarray*}
\end{mdframed}

\subsection{OCP Transcription}
Once the overall OCP has been defined in continuous time, it has to be effectively solved. To do so, it is necessary to cast it into discrete time, transcribing the OCP into an equivalent Non-Linear Programming (NLP) problem that can be solved by structure-exploiting solvers. In the present work, this is achieved with orthogonal collocation techniques \cite{orthogonal}, hence approximating the state trajectories with suitable $d^{th}$-order polynomial splines. The overall optimization horizon $T$ is broken down into $N$ intervals of equal length, and inside the generic interval $[t_k, \ t_{k+1}]$ we select $d$ Gauss-Legendre collocation points at which the dynamics as in (\ref{math: srbd_theory}) is enforced. In particular, we choose $2^{nd}$-order polynomials $\boldsymbol{\pi}_1(t) \in \mathbb{R}^4$ to represent the vertical trajectories of the feet, and let $\boldsymbol{u}_2$ free to vary at every collocation point in order to achieve a good shape of such trajectories (as shown in Fig. \ref{fig: goback_forces}). All the other state variables are approximated with $3^{rd}$-order polynomials $\boldsymbol{\pi}_2(t) \in \mathbb{R}^{24}$, keeping the control vector $\boldsymbol{u}_1$ constant throughout the whole interval. These choices grant an accurate approximation of the real dynamics while reducing the computational effort required. Also, the lifted method \cite{lifted} is used, integrating the system dynamics independently on each interval and enforcing continuity of the solution only at solution time.


\section{Experimental Validation}\label{sec:experiments}
The experimental validation has been performed on the CENTAURO platform \cite{kashiri2019centauro} powered by the XBotCore framework \cite{muratore2017xbotcore}. 
Different navigation patterns have been produced in order to highlight the variety of behaviours that can be generated by the proposed TO framework with minimum authoring effort. The transcribed OCP has been implemented in CasADi \cite{casadi} interfaced with Python, and solved employing the open-source solver Ipopt \cite{ipopt}. The performance of the algorithm on an Intel Core i7/1.8 GHz laptop for the different navigation patterns are shown in Table \ref{table: overview}. 
Once a solution for the OCP is produced, the optimal Cartesian references are fed to the CartesI/O framework \cite{laurenzi2019cartesi}, which relies on the hierarchical Inverse Kinematics library OpenSoT \cite{OpenSot17}, to track the planned trajectories. The considered SoT can be written using the Math of Task (MoT) formalism \cite{OpenSot17} as follows:
\begin{equation}
    \begin{pmatrix}
    \vspace{0.1cm}
    \left(\sum_i \   ^{\text{World}}\mathcal{T}_{{\text{Foot}}_i}^\text{[XYZ]}\right) \ / \\
    \vspace{0.1cm}
    \left(\sum_i   \mathcal{T}_{{\text{Wheel}}_i} + \sum_i \ ^{\text{World}}\mathcal{T}_{{\text{Ankle}}_i}^\text{[RPY]}\right)  \ / \\
    \vspace{0.1cm}
    ^{\text{World}}\mathcal{T}_{\text{Waist}} \ / \\
    \vspace{0.08cm}
    \mathcal{T}_{\text{Posture}}
    \end{pmatrix} << 
    \begin{pmatrix}
    \mathcal{C}_{\substack{\text{Joint.}\\\text{Lims}}}+\mathcal{C}_{\substack{\text{Vel.}\\\text{Lims}}}
    \end{pmatrix}
\label{math: sot_hybrid}
\end{equation}

The cost function weights remain constant in all the experiments performed. They have been tuned heuristically, to accommodate for the different orders of magnitude of the objectives' variables, as follows: $\gamma_{acc}=\gamma_{speed}=1$, $\gamma_{goal}=10$, $\gamma_{force}=5 \cdot 10^{-4}$, $\gamma_{feet}=20$, $\gamma_{yaw}=1.5$.

\subsection{Legged Locomotion} \label{sec: legged_forward}
Our framework can be used to produce pure legged locomotion by simply considering a lower order model discarding the wheel's steering angle, while constraining the foot sliding velocity. For this scenario, the cost function in \ref{sec:cost_funct} does not comprise the last two terms. Due to space limitation, the resulting behaviour is shown in the accompanying video, setting a goal 70 cm in front with 8 steps.

\subsection{Wheeled Navigation} \label{sec: pure_roll}
If no flight phase is allowed for the feet, the overall behaviour is similar to the one achievable with a car-like robot, as in \cite{laurenzi2019augmented}, but each wheel can be steered autonomously and their relative position can change. With a CoM goal that is placed 100 cm ahead and 60 cm to the left, and a final yaw orientation for the base set to be $90^{\circ}$ counterclockwise, a smooth wheeled motion is performed, see Fig. \ref{fig: results}.(a).

\subsection{Hybrid Wheeled-Legged Locomotion} \label{sec: hybrid_goback}
\begin{figure}
    \centering
    \includegraphics[width=.95\columnwidth]{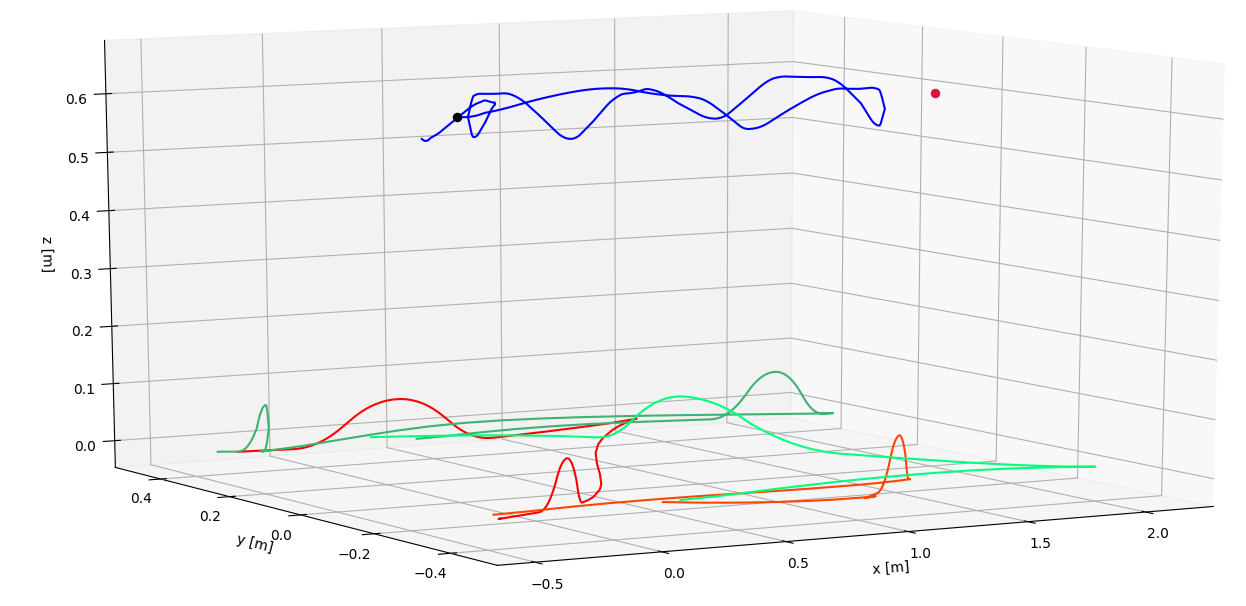}
    \caption{Results for \ref{sec: hybrid_goback} go\&back. The optimized trajectories for CoM (blue) and feet (green=front, red=hind) are shown, together with the initial position (black) and initial goal (red)}
    \label{fig: goback_3D}
\end{figure}

\begin{figure}
    \centering
    \includegraphics[width=.85\columnwidth]{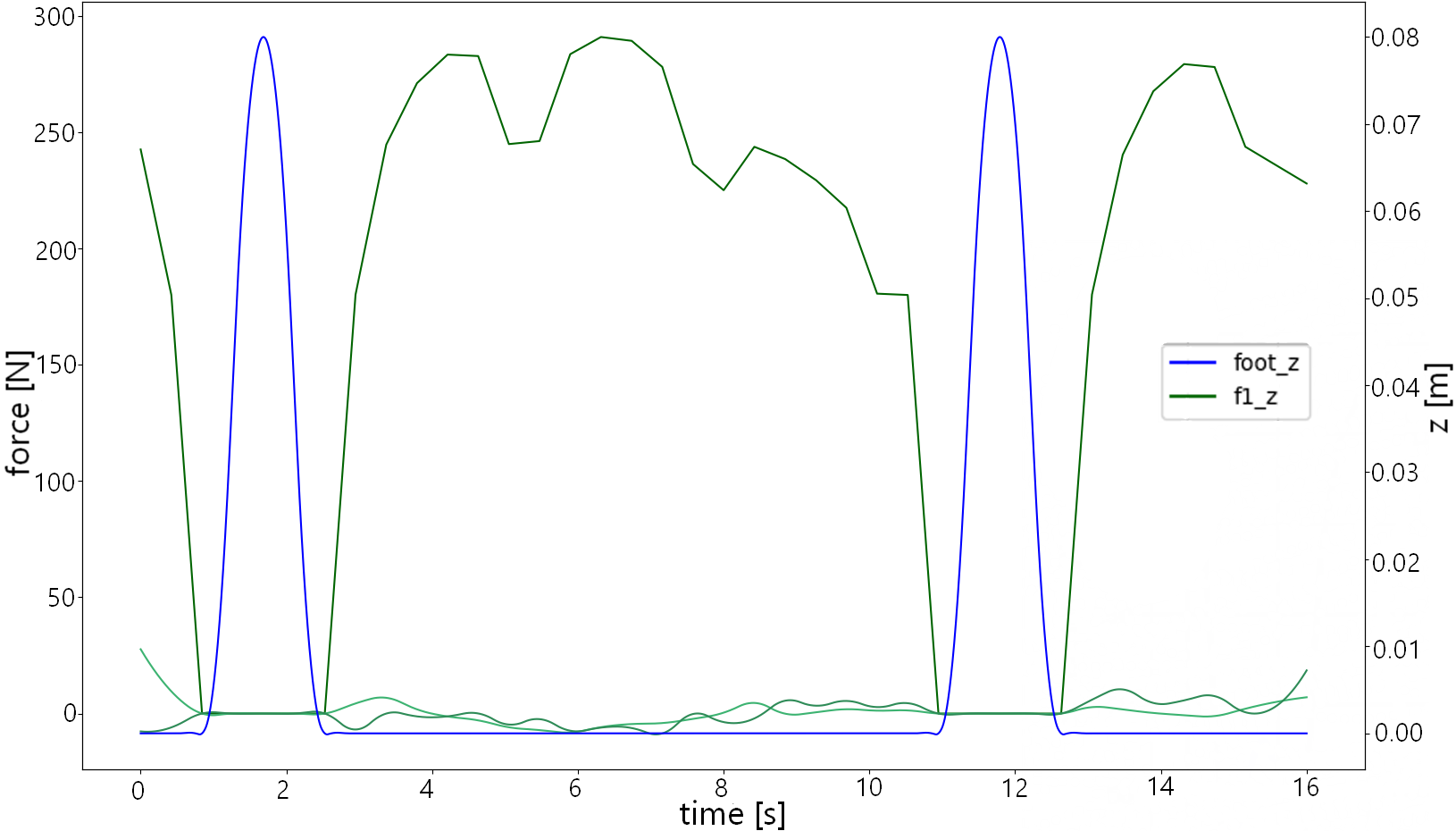}
    \caption{Results for \ref{sec: hybrid_goback} go\&back. Contact force in its 3D components is shown together with the vertical trajectory of the left hind foot.}
    \label{fig: goback_forces}
\end{figure}

\begin{figure}
    \centering
    \includegraphics[width=0.95\columnwidth]{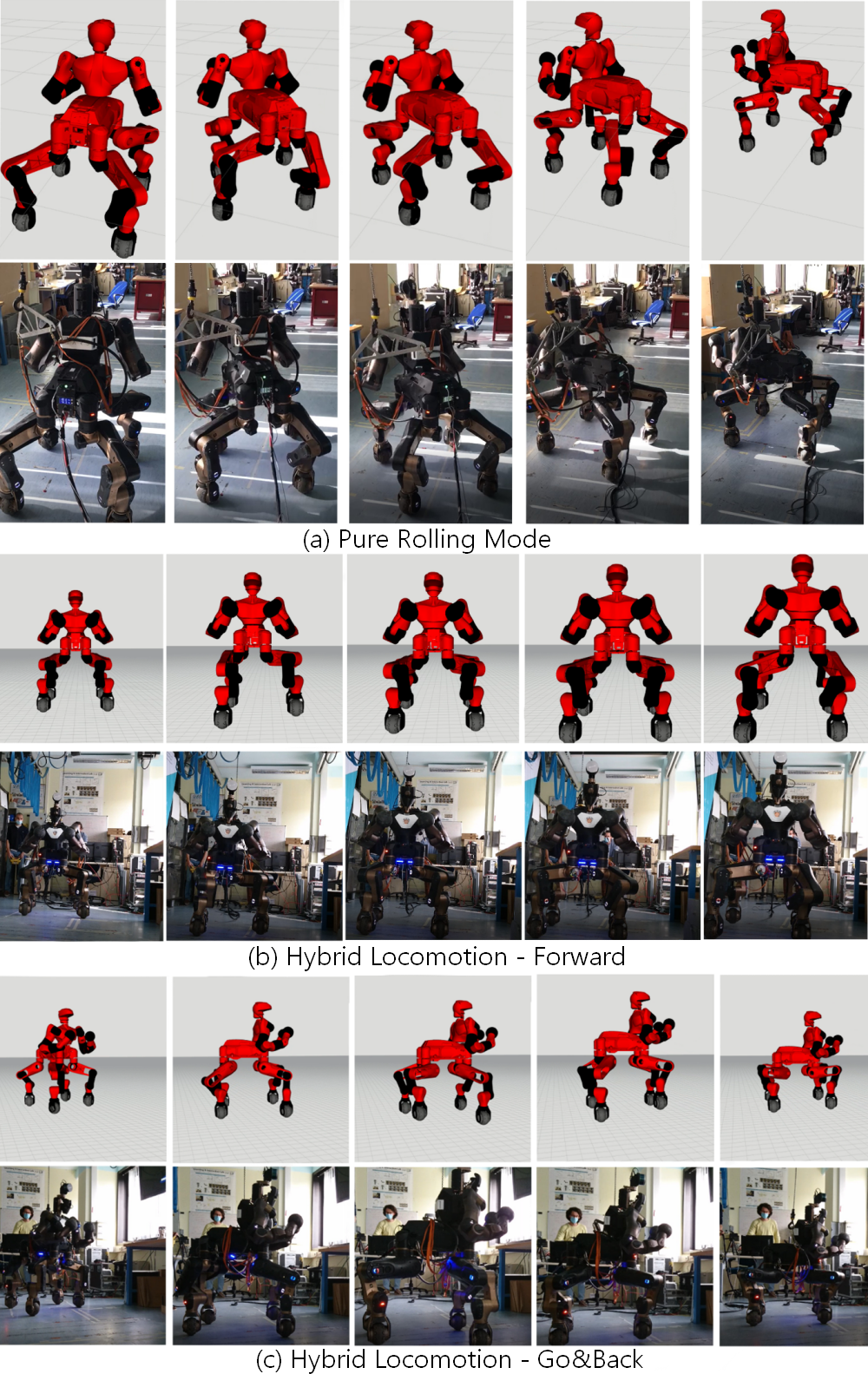}
    \caption{RViz visualization and experimental validation on CENTAURO of different locomotion gaits generated by the proposed TO framework.}
    \label{fig: results}
\end{figure}

Here the presented framework shows all its potential: forward hybrid locomotion as well as a hybrid \say{go\&back} behaviour is achieved. In the first case the robot moves to a goal 200 cm far, in the second case also the base orientation must change (by rotating $90^{\circ}$ counterclockwise) and halfway through the goal is switched so that the robot must go back to the origin while maintaining the same orientation. The behaviours achieved can be seen in Fig. \ref{fig: results}.(c). Fig. \ref{fig: goback_3D} and Fig. \ref{fig: goback_forces} show time histories of relevant quantities.
\begin{table}[h!]
\centering
\caption{Overview of the performances of the proposed algorithm 
}
\begin{adjustbox}{width=\columnwidth}
\begin{tabular}{c c c c c c} 
 \hline
Navigation type  & & &  prediction horizon & computation time & \#iterations \\
 \hline
 Legged  & & & 10s   & 19.0s & 37 \\ 
 Wheeled & & & 15s   & 20.0s & 21 \\
 Hybrid-forward & & & 10s   & 98.0s & 60 \\
 Hybrid-go\&back & & & 16s & 63.0 & 170 \\
 \hline
\end{tabular}
\end{adjustbox}
\label{table: overview}
\end{table}


\section{Conclusion and future works}\label{sec:conclusion}
In this paper, we presented the formulation, transcription and solution of an OCP capable of generating varied locomotion on the robot CENTAURO, focusing in particular on the integration of steerable wheels. Hybrid locomotion is achieved with the proposed approach, enabling the robot to perform fast and stable motions. The results have been experimentally validated on the real robot system, providing satisfactory results. 
As future works, reducing the TO computation time required would enable MPC approaches to the hybrid locomotion problem. Interesting extensions of our work include the automatic generation of the gait sequence, see \cite{wink3rdpaper}, providing the algorithm with the flexibility to optimize over the duration of each lift phase of the feet. Ultimately, sensor integration would enable to apply the proposed locomotion strategy e.g. in a gap crossing task.

\bibliographystyle{IEEEtran}	
\bibliography{biblio}

\end{document}